\documentclass[10pt,twocolumn,letterpaper]{article}

\usepackage{cvpr}
\usepackage{times}
\usepackage{epsfig}
\usepackage{graphicx}
\usepackage{amsmath}
\usepackage{amssymb}

\usepackage{makecell}
\usepackage{multirow}
\usepackage{bm}
\usepackage{placeins}
\usepackage{booktabs}

\usepackage{blindtext}
\usepackage{caption}
\newcommand{\R}{\mathbb{R}}
\newcommand\numberthis{\addtocounter{equation}{1}\tag{\theequation}}

\usepackage[pagebackref=true,breaklinks=true,letterpaper=true,colorlinks,bookmarks=false]{hyperref}

\cvprfinalcopy

\def\cvprPaperID{9109}

\ifcvprfinal\fi
\begin{document}

%%%%%%%%% TITLE
\title{Weakly-Supervised Mesh-Convolutional Hand Reconstruction in the Wild}

\author{Dominik Kulon$^{1, \;3}$ \hspace{1em}
	Riza Alp G\"uler $^{1, \;3}$ \vspace{0.15em} \\
	Iasonas Kokkinos $^{3}$  \hspace{1em}
    Michael Bronstein $^{1, \; 2, \;4}$\hspace{1em}
    Stefanos Zafeiriou $^{1, \;3}$ \vspace{0.3em}\\
	$^1$Imperial College London \hspace{1em} $^2$USI Lugano \hspace{1em} $^3$Ariel AI \hspace{1em} $^4$Twitter\\
	{\tt\scriptsize\{d.kulon17, r.guler, m.bronstein, s.zafeiriou\}@imperial.ac.uk 	}\hspace{1em}	
	{\tt\scriptsize iasonas@arielai.com}
}

\twocolumn[{
\maketitle
\vspace*{-10mm}
\begin{center}
    \centering
    \includegraphics[width=0.95\textwidth,trim={0 0 0 0},clip]{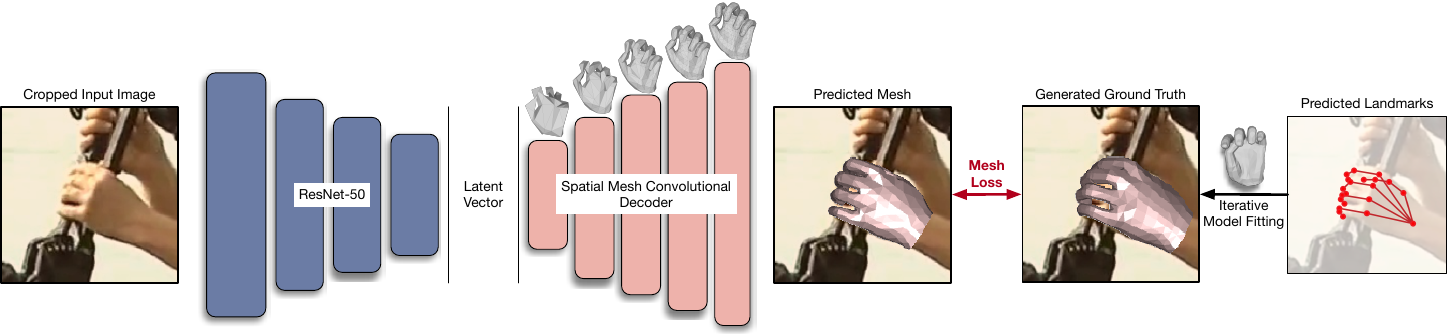}
    \captionof{figure}{
    We propose an approach for end-to-end neural network training with mesh supervision that is obtained through an automated data collection method. We process a large collection of YouTube videos and analyze them with 2D hand keypoint detector followed by parametric model fitting (right side). The fitting results are used as a supervisory signal (`mesh loss') for a feed-forward network with a mesh convolutional decoder tasked with recovering a 3D hand mesh at its output (left side).
    } 
    \label{fig:nn-diagram}
\end{center}
}]

%%%%%%%%% ABSTRACT
\begin{abstract}
We introduce a simple and effective network architecture for monocular 3D hand pose estimation consisting of 
an image encoder followed by a mesh convolutional decoder that is  trained through a direct 3D hand mesh reconstruction loss.
We train our network by gathering a large-scale dataset of hand action in YouTube videos and use it as a source of weak supervision. 
Our weakly-supervised mesh convolutions-based system largely outperforms state-of-the-art methods, even halving the errors on the in the wild benchmark. 
The dataset and additional resources are available at \url{https://arielai.com/mesh_hands}.
\end{abstract}

%%%%%%%%% BODY TEXT
\section{Introduction}
Monocular 3D hand reconstruction can facilitate a broad array of applications in human-computer interaction, augmented  reality, virtual telepresence, or sign language recognition. Still, 
the current state-of-the-art methods do not always generalize to samples captured in a non-laboratory environment. Our work is aimed at operating in the wild, and is therefore trained primarily on a large-scale dataset of hand action images gathered from YouTube videos; at the same time our system largely outperforms the current state-of-the-art as evaluated on public benchmarks. 

Even though our method can deliver a dense hand mesh, we outperform systems that try to solve a more narrow task 
of estimating coordinates of a sparse set of keypoints usually corresponding to hand joints and fingertips. The standard approach is to regress heatmaps of 2D landmarks and find a matching 3D pose \cite{DBLP:journals/corr/ZimmermannB17, 2017arXiv171203866P, 2017arXiv171201057M}. More recent works, that also focus on full-blown mesh reconstruction, rely on fitting deformable models to estimated landmarks, effectively exploiting a prior distribution in 3D shape space to perform 3D reconstruction \cite{2019arXiv190203451B, 2019arXiv190209305Z, 2019arXiv190404196B, Hasson:CVPR:2019}. Such models are either parameterized by angles of rotation and weights of a linear deformation function~\cite{MANO:SIGGRAPHASIA:2017} or latent features learned from  data~\cite{2019arXiv190300812G, dkulon2019rec}. We follow the latter approaches, which have proven to be the most promising, but we substantially improve their accuracy and robustness. 

Our first key contribution is a weakly-supervised training method for 3D mesh reconstruction. We introduce a fast and accurate method for dataset generation from unlabeled images, that relies on fitting a hand model to 2D keypoint detections while respecting a prior on rotation angles. Our data generation algorithm processes a sequence of YouTube videos, generating a curated dataset of images with 3D hand meshes. We have selected a filtered subset of 50,000 training annotations comprising hundreds of subjects performing a wide variety of tasks.

Beyond the data-driven advancements, we also obtain a substantial improvement by using a spatial mesh convolution method that is based on neighbourhood ordering. 
We train our end-to-end system from scratch
by using an objective function that merely enforces the reconstruction of a hand mesh that is aligned with the image. 

Our system attains 50\% lower hand pose estimation error compared to the currently best model on the in-the-wild scenario, which is our main focus. It also outperforms previous methods on hand pose estimation in controlled environments, without tuning or overfitting on a particular dataset.

To summarise, our contributions are as follows: 
\begin{itemize}
    \item We introduce an automated approach to generate training data from unannotated images for 3D hand reconstruction and pose estimation. The proposed method is scalable and does not require large computational resources.
    \item  We release the dataset of 50,000 meshes aligned with frames selected from over a hundred YouTube videos.
    \item We propose a simple loss function for mesh reconstruction that allows training neural networks with no intermediate supervision.
    \item We demonstrate that spatial mesh convolutions outperform spectral methods and SMPL-based~\cite{SMPL:2015, MANO:SIGGRAPHASIA:2017} models for hand reconstruction.
    \item We largely outperform the current state-of-the-art on pose estimation and mesh reconstruction tasks, while using a simple and  single-shot encoder-decoder architecture.
\end{itemize}

\section{Background}

We focus our attention on the hand pose estimation task using only RGB images as input. This is significantly more challenging than hand pose estimation when a depth sensor is also available~\cite{DBLP:journals/corr/abs-1712-03917, DBLP:journals/corr/SupancicRYSR15, ge2016robust, oberweger2017deepprior++, WetzlerBMVC15}. Hand mesh recovery is a more general task as it aims at reconstructing a relatively dense point cloud and  estimates both the pose and subject-specific shape of the target object. We categorize a method as belonging to the latter category if it allows generation of different subject-specific shapes.

\paragraph*{Hand Pose Estimation}

Multiview bootstrapping \cite{DBLP:journals/corr/SimonJMS17} is an iterative improvement technique where the initial annotations of a hand from multiple views, obtained by executing a 2D keypoint detector, are triangulated in 3D and reprojected for training in the next iteration if they meet the quality criteria. This system is part of the OpenPose framework~\cite{DBLP:journals/corr/WeiRKS16} that we use for weak supervision.

Numerous approaches have been proposed to regress a 3D pose from estimated 2D keypoints~\cite{DBLP:journals/corr/ZimmermannB17, 2017arXiv171203866P, 8621059, cai2018_weakly}. Dibra et al.~\cite{Dibra_2018_CVPR_Workshops} incorporate depth regularization to supervise 3D pose estimation represented by angles of rotation. In contrast, variational approaches learn a shared latent space by an alternated training between variational autoencoders \cite{spurr2018cvpr}, disentanglement of the latent representation \cite{yang2018disentangling}, or the alignment of latent spaces~\cite{yang2019iccv}. This allows for sampling outputs in any of the learned modalities. Gao et al.~\cite{gao2019object} address the problem of estimating the pose of hands interacting with objects in a variational framework. Generative networks can also be used to transfer synthetic images into a space of real-world images \cite{2017arXiv171201057M} by modifying a cycle adversarial loss~\cite{DBLP:journals/corr/ZhuPIE17} to include a geometric consistency term that ensures pose-preservation during an image translation. Tekin et al.~\cite{tekin2019h+} estimate hand and object poses with a single-shot network.

Recently, Cai et al.~\cite{cai2019exploiting} proposed to build a spatio-temporal graph and use graph convolutions in the spectral domain to estimate a 3D pose.

\paragraph*{Hand Mesh Recovery} 

MANO~\cite{MANO:SIGGRAPHASIA:2017} is a hand model parameterized by angles of rotations specified for each joint in the kinematic tree and blend weights of a linear function that models the shape of the person. Different articulations are obtained by applying linear blend skinning which interpolates rotations matrices assigned to the joints to transform a vertex according to the angles. Additionally, pose-dependent corrective offsets are learned to address the loss of volume caused by the skinning method. Recently, numerous works have been proposed to find a 3D pose by regressing MANO and camera parameters.

Boukhayma et al.~\cite{2019arXiv190203451B} regress these parameters from an image and heatmaps obtained from OpenPose. The method is evaluated on images in the wild but the error is relatively high as we show in the Evaluation section. The works by Zhang et al.~\cite{2019arXiv190209305Z} and Baek et al.~\cite{2019arXiv190404196B} introduce methods that iteratively regress model parameters from the heatmaps. Hasson et al.~\cite{Hasson:CVPR:2019} predict MANO parameters and reconstruct objects the hands interact with.

Ge et al.~\cite{2019arXiv190300812G} use graph convolutions in the spectral domain to recover a hand mesh. The system includes a heatmap regression module and, in case of real-world images, is supervised with 2D keypoints, depth maps, and meshes approximated with a pretrained model from ground truth heat maps. Kulon et al.~\cite{dkulon2019rec} reconstruct a hand from an image by regressing a latent vector of a pretrained mesh generator and estimating a camera projection. 

In this paper, we introduce an architecturally simpler approach that significantly outperforms prior works by computing loss on meshes with points localized in the image coordinate system and leveraging spatial mesh convolutions.

\paragraph*{Body Mesh Recovery} 

Approaches based on regressing MANO parameters described in the previous subsection originate from works on body mesh recovery~\cite{Bogo:ECCV:2016, DBLP:journals/corr/Lassner0KBBG17, hmr_kanazawa, tan2017indirect, 2018arXiv180504092P, 2018arXiv180404875V, SMPL-X:2019, DBLP:journals/corr/abs-1808-05942, DBLP:journals/corr/abs-1801-01615, DBLP:journals/corr/abs-1812-01598, SPIN:ICCV:2019}.
Our work is similar to Kolotouros et al \cite{Kolotouros_2019_CVPR}, who  encode an image and concatenate its features with coordinates of a template mesh in a canonical pose. The resulting graph is passed through a sequence of spectral graph convolutions to recover the body model corresponding to the image. In contrary, we decode a mesh directly from the image encoding and apply spatial convolutions with pooling layers.

\paragraph*{Geometric Deep Learning} 

Our interest lies in applying deep learning to learn invariant shape features in a triangular mesh~\cite{bronstein2017geometric}. To this end, spectral approaches~\cite{DBLP:journals/corr/BrunaZSL13, DBLP:journals/corr/DefferrardBV16, kipf2017semi, DBLP:journals/corr/LevieMBB17} express convolutions in the frequency domain. Spatial convolutions define local charting \cite{DBLP:journals/corr/MasciBBV15, DBLP:journals/corr/BoscainiMRB16, DBLP:journals/corr/MontiBMRSB16} that seems more suitable for learning on manifolds or designing meaningful pooling functions. Alternatively, convolution operators can be defined with an attention mechanism to weight the neighbour selection~\cite{DBLP:journals/corr/VermaBV17, velikovi2017graph}.

In terms of applications similar to our method, Ranjan et al.~\cite{DBLP:journals/corr/abs-1807-10267} use fast spectral convolutions~\cite{DBLP:journals/corr/DefferrardBV16} to find a low-dimensional non-linear representation of the human face in an unsupervised manner. Kulon et al.~\cite{dkulon2019rec} show that autoencoders can be used to learn a latent representation of 3D hands. Recently, a spiral operator \cite{DBLP:journals/corr/abs-1809-06664} has been incorporated into a convolutional framework to train a deformable model of the human body~\cite{2019arXiv190502876B} or solve mesh correspondence and classification tasks~\cite{gong2019spiralnet++}. In this paper, we apply spiral filters to generate hands directly from an image encoding.

\section{YouTube Hands-in-the-Wild Dataset} \label{section:data-collection}

Despite the wealth of possible applications, 
there are no datasets for monocular 3D hand reconstruction from RGB images in-the-wild. The only existing collection of images of hands captured in a non-laboratory environment \cite{DBLP:journals/corr/SimonJMS17} contains less than 3000 samples with a manual annotation of 2D points. 

In order to train a neural network for the 3D reconstruction of hands across different domains, we have built a system for fast and automated dataset generation from YouTube videos, providing us with a diverse set of challenging hand images. Instead of annotating them manually, we use a weakly-supervised approach that first detects keypoints using OpenPose, and then lifts them into 3D shapes by iteratively fitting a 3D deformable model. 
While the accuracy of the trained model could be bounded by the performance of OpenPose, we show that it results in a state-of-the-art 3D hand reconstruction and pose estimation system when evaluated on external datasets with manual annotations.

\subsection{3D Shape Representation}
Our approach relies on fitting the MANO model in tandem with a prior on angles of rotation. 
MANO predicts $N = 778$ vertices on the hand surface and $K = 16$ joints through a differentiable function $M(\beta, \theta)$ that maps shape ${\beta} \in \R^{|\beta|}$ and pose ${\theta} \in \R^{K \times 3}$ parameters into an instance of the model represented by a mesh:
\begin{equation}
M(\beta, \theta, \vec T_{\delta}, s; \phi) : \R^{|\beta| \times |\theta| \times |\vec T_{\delta}| \times |s|} \rightarrow \R^{N \times 3}
\end{equation}
where $|\beta|$ depends on the training procedure and $\phi$ is a set of learned model parameters that we omit in a further discussion. Moreover, we have camera parameters $s$ for scaling the model and $\vec T_{\delta} \in \R^3 $ to translate it. Global orientation is modeled by the first row of $\theta$. 

Rather than model the joint angles as free variables, 
which can lead to impossible estimates during optimization, we constrain them to lie in the convex hull of some pre-computed cluster centers. For  joint $i$ we use $C = 64$ Euler-angle clusters $P_i^1, ..., P_i^{C}$ obtained via k-means~\cite{Guler_2019_CVPR} and express any angle as follows: 
\begin{equation}
\theta_i =  P(w)_i = \frac{\sum_{c = 1}^{C} \exp(w^{c}) P_i^c}{\sum_{c = 1}^{C} \exp(w^{c}) }.
\end{equation}
While constraining the final angle estimate to take  plausible values, at the same time this expression allows us to optimize over unconstrained variables $w^{c}$. We represent all constrained angles in terms of a parameter matrix $w \in \R^{K \times C}$, 
while allowing global orientation $w_0$ to be unrestricted. 

This simple approach requires only a small dataset of angles, unlike VAE priors; restricts pose space to plausible poses, unlike PCA priors that are characterized by unrealistic interpolation; and allows fitting unseen poses. However, it does not model pairwise dependencies, which we leave to future work.

\subsection{Parametric Model Fitting} \label{section:fitting}
Our supervision comes in the form of 2D landmarks extracted by OpenPose. We define a fitting procedure that tries to find a matching 3D pose from the MANO mesh through a sparse matrix $\mathcal{J} \in \R^{N \times (K + F)}$ that regresses from model vertices to $K=16$ joint and $F = 5$ fingertip positions, delivering the hand pose $J \in \R^{(K + F) \times 3}$:
\begin{equation}
J(\beta, w, \vec T_{\delta}, s) =  \mathcal{J}^T M(\beta, P(w), \vec T_{\delta}, s).
\label{eq:joints-regressor}
\end{equation}

We fit the model to 2D annotations by minimizing the following objective:
\begin{equation}
\{\beta^{*}, w^{*}, \vec T_{\delta}^{*}, s^{*}\} = \arg\min_{\beta, w, \vec T_{\delta}, s} (E_{2D} + E_{bone} + E_{reg}),
\end{equation}
consisting of a 2D reprojection term $E_{2D}$, 
a bone length preservation cost $E_{bone}$, and a regularization term $E_{reg}$.

In particular, the joint error term minimizes the distance between 2D joints
\begin{equation}
E_{2D}(\beta, w, \vec T_{\delta}, s) = ||\Lambda_{2D} (\Pi_K(J(\beta, P(w), \vec T_{\delta}, s)) - Y)||^2
\label{eq:joints-error}
\end{equation}
where $Y$ are 2D detector predictions and
$\Pi_K$ is the intrinsic camera projection to 2D, initialized as in the approach of Pavlakos et al.~\cite{SMPL-X:2019}. $\Lambda_{2D}$ is an experimentally chosen mask that amplifies the influence of fingertips by $1.7$ and wrist by $2.5$ and reduces influence of the metacarpophalangeal (MCP) joints (base of each finger) that are often inaccurately annotated by $0.7$. 

The bone loss  $E_{bone}$ ensures that the length of each edge in the hand skeleton tree $\mathcal{E}$ is preserved, i.e.
\begin{equation}
    E_{bone}(\beta, w, \vec T_{\delta}, s)  = \sum_{(i, j) \in \mathcal{E}} | \; ||J_{2D_j} - J_{2D_i}|| - ||Y_{j} - Y_{i}|| \; |
\end{equation}
where $J_{2D_i} = \Pi_K(J(\beta, w, \vec T_{\delta}, s))_i$.

The regularization term $E_{reg}(\beta, \theta) = \lambda_{\theta} ||\theta||^2 + \lambda_{\beta} ||\beta||^2$ penalizes deviations from the mean pose to ensure realistic deformations. The hyperparameters $\lambda_{\theta} = 0.1$ and $\lambda_{\beta} = 1000$ were chosen experimentally.
\paragraph{Optimization}
We use the Adam optimizer with different learning rates for camera, pose, and shape parameters ($10^{-2}$, $10^{-2}$, $10^{-5}$ respectively) and  small learning rate decay (multiplicative factor of $0.95$) after each 500 iterations. We start with 1,500 iterations optimizing over camera parameters and global orientation where the joints set is reduced to a wrist and MCP joints excluding thumb. Afterwards, we perform 2,500 iterations over all parameters. We fit 4,000 samples per batch on GeForce RTX 2080 Ti which takes on average 10 min.

\subsection{Automated Data Collection} \label{section:automated}

The data collection system iterates a list of YouTube links, downloads a video, extracts frames, runs OpenPose, fits MANO to each frame, and selects a small subset of filtered samples. The depth of the projected mesh is proportional to the ratio of standard deviations between $X$ coordinates of the projected mesh and its world position. 

Filtering of plausible samples is performed by thresholding total OpenPose confidence score, per-joint confidence score, and the mean squared error between projected MANO joints and OpenPose predictions normalized by the distance from the camera.

To create our \texttt{YouTube} dataset, we process 102 videos for the training set and randomly select at most 500 samples per video that meet the threshold conditions. Most of the samples cover sign language conversations performed by people of wide variety of nationality and ethnicity. Some videos include, for example, a hundred people signing a common phrase to a webcam all over the world. The validation and test sets cover 7 videos with an empty intersection of subjects with the training set. We selected test videos to be diverse and challenging including conversations captured outdoor, dance poses, American, Australian, and Taiwanese sign languages (Figure~\ref{fig:qualit}, left half). Additionally, we run our system on the COCO dataset~\cite{2014arXiv1405.0312L} and extract 7,048 hand images for training. The combined training set contains 54,173 samples and validation and test sets count 1,525 images each. 

\section{Hand Reconstruction Network}

We propose a simple encoder-decoder system, as demonstrated in (Figure~\ref{fig:nn-diagram}), that directly reconstructs the mesh in image coordinates. We use a spatial convolutional mesh decoder $\mathcal{D}_{mesh}$, which we experimentally have shown to be superior with respect to alternative decoding strategies. In the following subsections we explain the spatial operator of choice and the upsampling method which constitute the building blocks of our decoder. 

\begin{figure}[h]
\centering
  \centering
  \includegraphics[width=\linewidth]{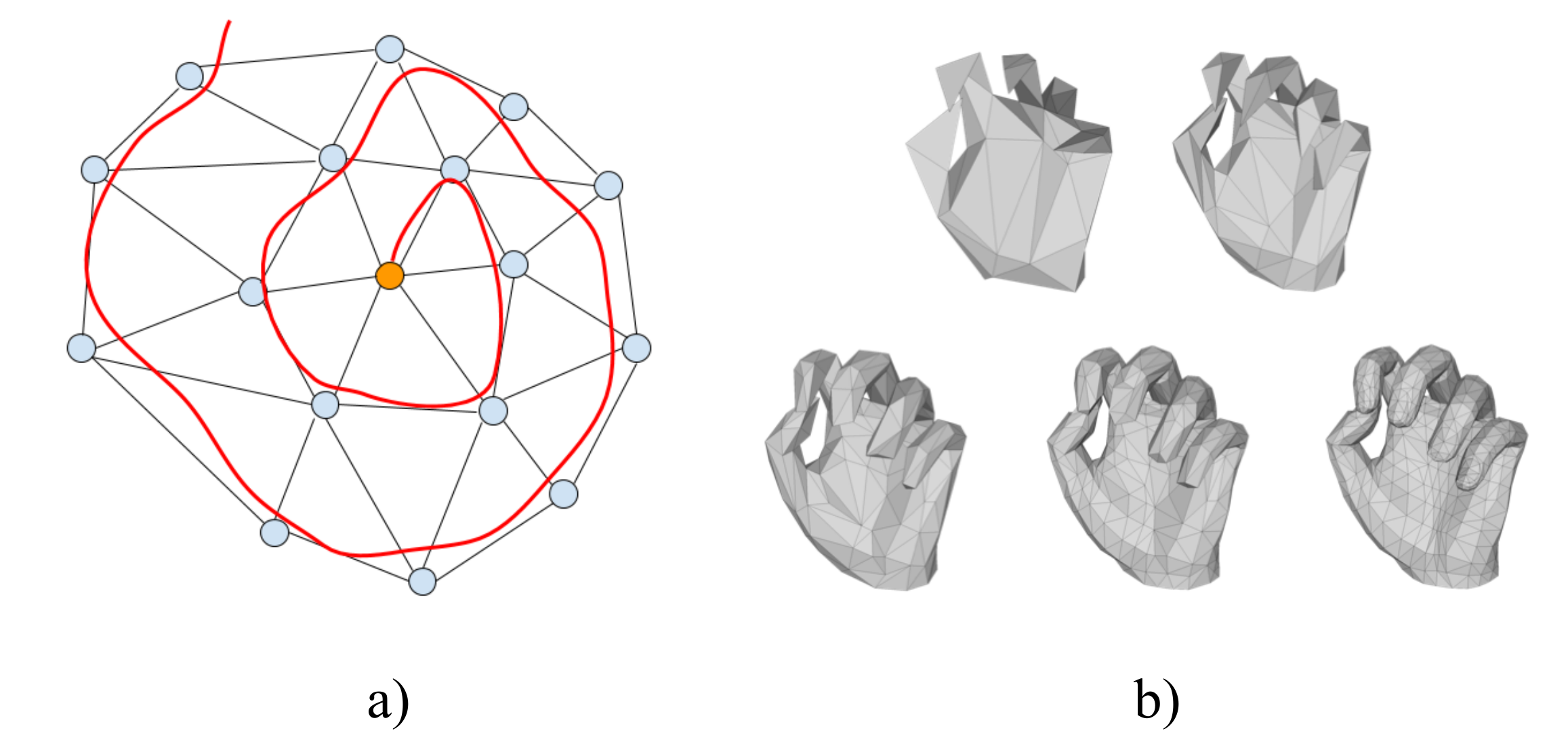}
\captionof{figure}{ a) Spiral selection of neighbours for a center vertex in clockwise order (taken from 
\cite{2019arXiv190502876B}). b) The hierarchy of topologies used in our decoder.}
  \label{fig:spiral-neighbour}
  \label{fig:upsampling}
\end{figure}

\subsection{Spiral Operator}

In our decoder we use the spiral patch operator for constructing spatial neighborhoods. Lim et al.~\cite{DBLP:journals/corr/abs-1809-06664} define a spiral selection of neighbours for a center vertex $v$ by imposing the order on elements of the $k$-disk obtained in the following way:
\begin{align*}
    0\text{-ring}(v) &= \{v\}, \\
    (k\!+\!1)\text{-ring}(v) &= N(k\text{-ring}(v)) \,\backslash\, k\text{-disk}(v),\\
    k\text{-disk}(v) &= \cup_{i = 0 \dots k}\, i\text{-ring}(v), \numberthis
\end{align*}
where $N(V)$ is the set of all vertices adjacent to any vertex in the set $V$. The spiral patch operator~\cite{2019arXiv190502876B} is then the ordered sequence $S(v)$
\begin{align*}
    \text{S}(v) = (v, 1\text{-ring}(v), \dots,  k\text{-ring}(v)). \numberthis
\end{align*}
As orientation, we follow the clockwise direction from each vertex and randomly initialize the first neighbour of $v$. An example spatial configuration is demonstrated in Figure~\ref{fig:spiral-neighbour}. The spiral convolution of features $f({S}(v))$ with the kernel $g_\ell$ can then be defined by
\begin{equation}
    (f * g)_v = \sum_{\ell=1}^{L} g_\ell \, f\big(S_\ell(v)\big),
\end{equation}
where the spiral length $L$ is fixed for each vertex in the same layer of a neural network.

\subsection{Sampling}

We create a hierarchy of meshes with the number of vertices reduced by the factor of 2 at each stage of the decoder. We contract vertex pairs based on quadric error metrics~\cite{garland1997surface} 
that allow the coarse vertices to be a subset of the denser mesh. We project each collapsed node into the closest triangle in the downsampled mesh \cite{DBLP:journals/corr/abs-1807-10267} and use the barycentric coordinates of the projected vertex to define interpolation weights for the upsampling matrix.

More specifically, we project a vertex $v_q \in V$ discarded during contraction into the closest triangle $v_i, v_j, v_k \in V_d$ of the downsampled mesh $V_d$ obtaining $\tilde{v}_p$. We compute its barycentric coordinates $w_1, w_2, w_3$ such that $\tilde{v}_p = w_1 v_i + w_2 v_j + w_3 v_k$ and $w_1 + w_2 + w_3 = 1$. The upsampling matrix $Q_u \in \R^{m \times n}$, for which $V_u = Q_u V_d$ and $m > n$ hold, is formed by setting $Q_u(q,i) = w_i$, $Q_u(q,j) = w_j$, $Q_u(q,k) = w_k$, and $Q_u(q,l) = 0$ for $l \not\in \{i, j, k\}$.

The resulting topology hierarchy obtained using the sampling strategy is shown in Figure~\ref{fig:upsampling}. The number of vertices at each layer $n \in \{51, 100, 197, 392, 778\}$.

\subsection{Architecture}

\begin{figure}[h]
\centering
  \centering
  \includegraphics[width=\linewidth]{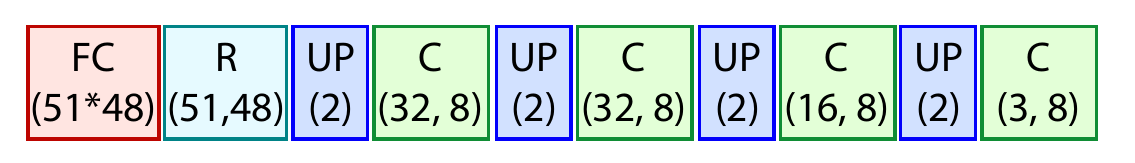}
\captionof{figure}{ Architecture of the spiral decoder. $C(w, l)$: A spiral convolution with $w$ filters and a spiral length $l$; $U(p)$: upsampling by a factor $p$; $FC(d)$: a fully connected layer with $d$ output dimension; $R(d_1, d_2)$: vector reshaping into a $d_1 \times d_2$ matrix.}
  \label{fig:architecture}
\end{figure}

Given an image crop $X$ centered around the hand, we embed it into a latent vector $Z = {E}_{image}(X)$ with 64 parameters. The decoder, $\mathcal{D}$, takes the embedding as input and produces a mesh $\mathcal{Y} = \mathcal{D}_{mesh}(Z)$.  We use a standard ResNet-50 network as the encoder, ${E}_{image}$. The architecture of the spiral decoder is detailed in Figure~\ref{fig:architecture}. If the spiral sequence is shorter than required due to erroneous triangulation or mesh border, we pad it with a node initially centered at 0. We only consider $k\text{-disks}$ for $k=2$. We choose leaky ReLU for the activation function based on experimental evaluation. 

\subsection{Training}

The loss function consists of the L1 vertex reconstruction term and edge length preservation component
\begin{equation}
\begin{split}
\mathcal{L} & =
 \lambda_{vertex} |\hat{\mathcal{Y}} - \mathcal{Y}|_1 \\ 
 & + \lambda_{edge} \sum_{(u, v) \in \mathcal{E}_{mesh}} | \; ||\hat{\mathcal{Y}}_v - \hat{\mathcal{Y}}_u|| - ||\mathcal{Y}_{v} - \mathcal{Y}_{u}|| \; |
\end{split}
\end{equation}
for ground truth meshes $\hat{\mathcal{Y}}$ and set of edges $\mathcal{E}_{mesh}$. Hyperparameters are set to $\lambda_{vertex} = 0.01$ and $\lambda_{edge} = 0.01$. There is no explicit pose estimation loss because we found it to have no effect. The joint coordinates are obtained during evaluation from a mesh as in Equation~\ref{eq:joints-regressor}.  

The network is trained with the Adam optimizer with learning rate $10^{-4}$ for 150 epochs. Learning rate decay with factor $0.1$ occurs after 90-th and 120-th epoch. The images are normalized with the mean and standard deviation computed from the ImageNet training set and output meshes are normalized based on statistics computed from the subset of our training data. We augment the data with random image crops and transformations. The data augmentation is necessary for generalization to real world examples where the input images cannot be cropped based on the ground truth annotations. We train the network for 2 days on a single GeForce RTX 2080 Ti with a batch size 32 and crop size of $192 \times 192$.

\section{Evaluation} 

Mesh reconstruction methods have started gaining popularity only recently and therefore there are no well-established benchmarks for hand recovery. To show the robustness of our method, we evaluate it on popular hand pose estimation datasets. Moreover, we show the hand reconstruction performance on the \texttt{FreiHAND} dataset and conduct a self-comparison study of different mesh decoders on our \texttt{YouTube} dataset. Lastly, we show that both major contributions, neural network architecture and data annotation system, result in the state-of-the-art performance in the absence of one or other.

\subsection{Datasets}

The \texttt{MPII+NZSL} (\texttt{MPII}) dataset \cite{DBLP:journals/corr/SimonJMS17} contains in the wild images collected by manually annotating two publicly available datasets with 2D landmarks: the MPII Human Pose dataset \cite{6909866} showing every-day human activities and a set of images from the New Zealand Sign Language Exercises \cite{nzsl}. It includes blurry, low-resolution, occluded, and interacting with objects hand images what makes the dataset particularly challenging. Training and test sets count 1,912 and 846 samples.

The \texttt{Rendered Hand Pose Dataset} (\texttt{RHD}) consists of 41,258 training and 2,728 testing samples of rendered characters~\cite{DBLP:journals/corr/ZimmermannB17}. It has been commonly used for RGB-based hand pose estimation due to its large size and challenging viewpoints.

\texttt{FreiHAND} is a recently released dataset with 130,240 training images~\cite{Freihand2019}. It is the only dataset that contains 3D mesh annotations with backgrounds artificially blended in place of the green screen. The test set consisting of 3,960 samples was collected without the green screen in a controlled outdoor and office environment. It contains difficult poses with object interactions and varying lighting. The test set annotations are not available and evaluation is performed by anonymously submitting predictions to the online competition. 

\texttt{Monocular Total Capture} (\texttt{MTC}) is a recent human motion dataset containing body, hands, and face annotations of 40 subjects~\cite{DBLP:journals/corr/abs-1812-01598}. The dataset was collected in a laboratory environment posing a risk of domain overfitting; however, due to its diverse and mostly accurate annotations it can be used to learn a robust prior on 3D hand shapes. We select a subset of 30,000 images for training filtered to reduce similar and mean poses.

\texttt{Stereo Hand Pose Tracking Benchmark} (\texttt{STB}) contains 15,000 training images of a single subject performing random poses and counting gestures with 5 different backgrounds and a test set of 3,000 images with the same background~\cite{DBLP:journals/corr/ZhangJCQXY16}. The dataset has been solved in recent publications~\cite{2019arXiv190300812G} with a high error tolerance due to inaccurate ground truth annotations.

Finally, our \texttt{YouTube} dataset is described in Section~\ref{section:automated}.

For each of the datasets, beside \texttt{FreiHAND} that provides vertex annotations, we fit the MANO model following the optimization approach described in Section~\ref{section:fitting}. In case of 3D annotations, we only change the intrinsic camera to the identity function in Equation~\ref{eq:joints-error} and then we project the output mesh based on dataset camera parameters. The depth is reconstructed as before.

\subsection{Metrics}

To evaluate hand pose estimation and mesh reconstruction performance, we measure the average euclidean distance (Pose/Mesh Error), the percentage of correct points (2D/3D PCK) for different thresholds, and the Area Under Curve (AUC) for PCK. For 3D benchmarks, we measure the error after rigid alignment. Before computing a 2D error, we orthographically project the estimated 3D pose. For self-comparison study, we report the mean absolute error (MAE) in addition to Mesh Error.

\noindent\begin{minipage}{\linewidth}
    \centering
    
    \includegraphics[width=1\linewidth]{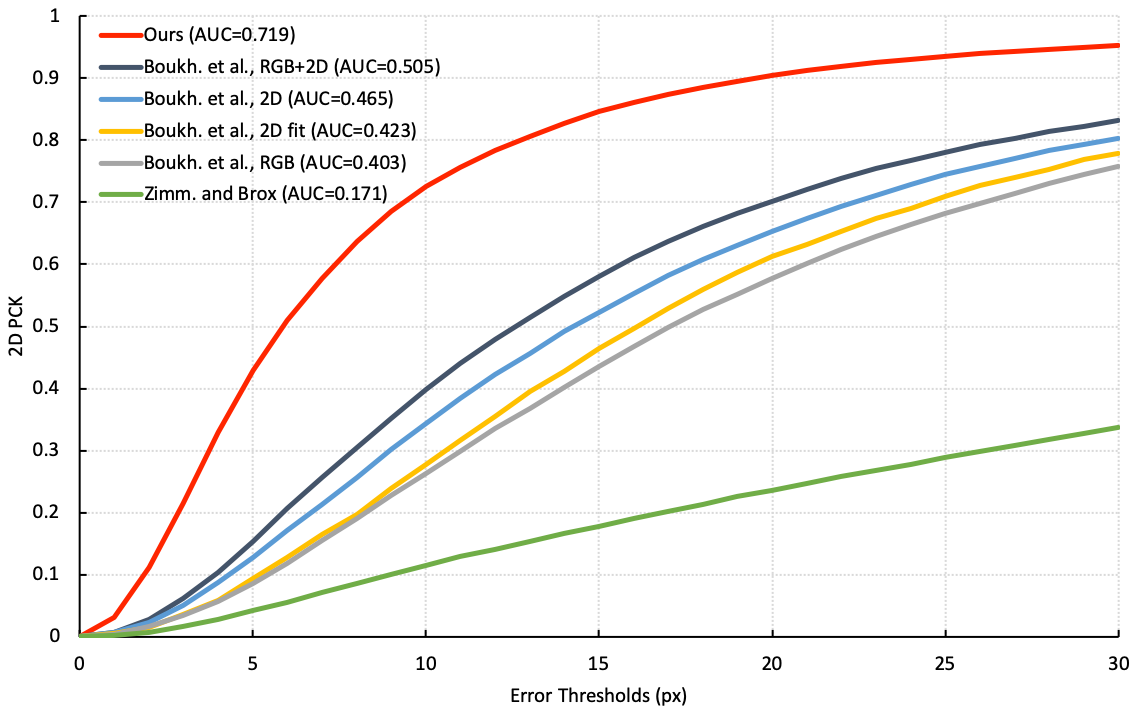}
    \par\bigskip
       \resizebox{\columnwidth}{!}{%
    \begin{tabular}{|| c || cccccc ||}
    \hline
        & Ours  &  
        \begin{tabular}[c]{@{}c@{}}Boukh. et al. \\ RGB+2D~\cite{2019arXiv190203451B}\end{tabular} &
        \begin{tabular}[c]{@{}c@{}}Boukh. et al. \\ RGB~\cite{2019arXiv190203451B}\end{tabular} & 
        \begin{tabular}[c]{@{}c@{}}Boukh. et al. \\ 2D Fit~\cite{2019arXiv190203451B}\end{tabular} & 
        \begin{tabular}[c]{@{}c@{}}Boukh. et al. \\ RGB~\cite{2019arXiv190203451B}\end{tabular} & Z\&B~\cite{DBLP:journals/corr/ZimmermannB17} \\ [0.5ex] 
     \hline\hline
    Pose Error [px] & \textbf{9.27} & 18.95                & 22.36            & 20.65                & 23.04             & 59.4        \\
    \hline
    \end{tabular}
    }
    \captionof{figure}{2D PCK and average 2D joint distance (px) to ground-truth for \texttt{MPII}.}
    \par\bigskip

    \label{fig:mpii-results}
\end{minipage} 

\noindent\begin{minipage}{\linewidth}
    \centering
    
    \includegraphics[width=1\linewidth]{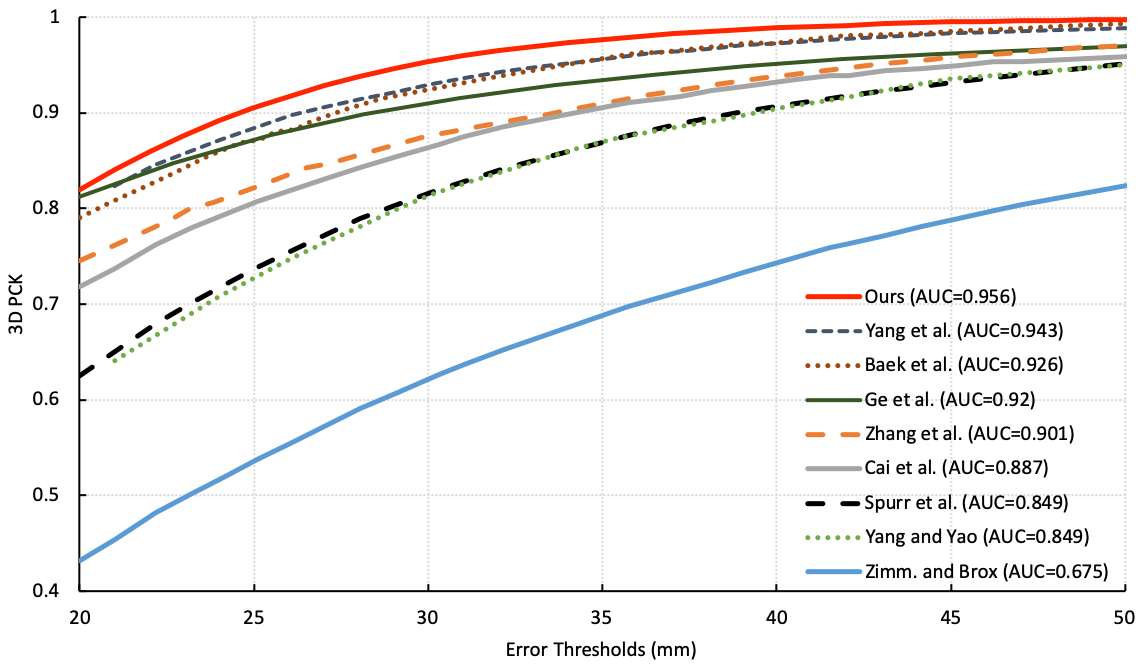}
    \par\bigskip
       \resizebox{\columnwidth}{!}{%
\begin{tabular}{||c||ccccc||}
\hline
    & Ours   & Yang et al.\cite{yang2019iccv} & Spurr et al.~\cite{spurr2018cvpr} & Y\&Y~\cite{yang2018disentangling} & Z\&B~\cite{DBLP:journals/corr/ZimmermannB17} \\ [0.5ex] 
 \hline\hline
Pose Error [mm] & \textbf{10.92} & 13.14       & 19.73        & 19.95        & 30.42         \\
\hline
\end{tabular}
    }
    \captionof{figure}{3D PCK and average 3D joint distance (mm) to ground-truth for \texttt{RHD}.}

    \label{fig:rhd-results}
\end{minipage} 

\subsection{Hand Pose Estimation}

Figure~\ref{fig:mpii-results} shows evaluation results on the \texttt{MPII} dataset. Our system significantly outperforms other approaches, halving the Pose Error of the leading MANO-based method and improving AUC by 0.21 points. 

Figure~\ref{fig:rhd-results} shows 3D pose estimation results on \texttt{RHD}. This is a very popular benchmark on which we also outcompete existing methods by a significant margin.

The proposed system also achieves a superior performance on the \texttt{FreiHAND} dataset as proved in Table~\ref{table:freihand-eval}. Notably, most of the competing methods in this benchmark rely on MANO parameters regression.

\begin{table*}[]
\centering
  \scalebox{0.9}{

\begin{tabular}{||l||llllll||}
\hline
         & Ours           & MANO Fit~\cite{Freihand2019} & MANO CNN~\cite{Freihand2019} & Boukh. et al.~\cite{2019arXiv190203451B} & Hasson et al.~\cite{Hasson:CVPR:2019} & Mean Shape~\cite{Freihand2019} \\ [0.5ex] 
 \hline\hline
Pose Error [mm] $\downarrow$   & \textbf{0.84}  & 1.37     & 1.1      & 3.5           & 1.33          & 1.71       \\
Mesh Error [mm] $\downarrow$    & \textbf{0.86}  & 1.37     & 1.09     & 1.32          & 1.33          & 1.64       \\
\hline \hline
Pose AUC $\uparrow$ & \textbf{0.834} & 0.73     & 0.783    & 0.351         & 0.737         & 0.662      \\
Mesh AUC $\uparrow$ & \textbf{0.83}  & 0.729    & 0.783    & 0.738         & 0.736         & 0.674      \\
\hline \hline
F@5 [mm] $\uparrow$       & \textbf{0.614} & 0.439    & 0.516    & 0.427         & 0.429         & 0.336      \\
F@15 [mm] $\uparrow$       & \textbf{0.966} & 0.892    & 0.934    & 0.894         & 0.907         & 0.837      \\
\hline
\end{tabular}
}
\caption{3D pose estimation and mesh reconstruction performance on \texttt{FreiHAND}. For the first two rows lower is better, for the other rows higher is better.}
\label{table:freihand-eval}
\end{table*}

\subsection{Mesh Reconstruction}

We evaluate the quality of mesh reconstructions on the FreiHAND dataset (Table~\ref{table:freihand-eval}). In addition to the standard metrics, we report the F-score at a given threshold $d$ ($F@d$) which is the harmonic mean of precision and recall~\cite{Knapitsch:2017:TTB:3072959.3073599}. 

\begin{figure*}[t]
    \centering
    \includegraphics[width=1\linewidth]{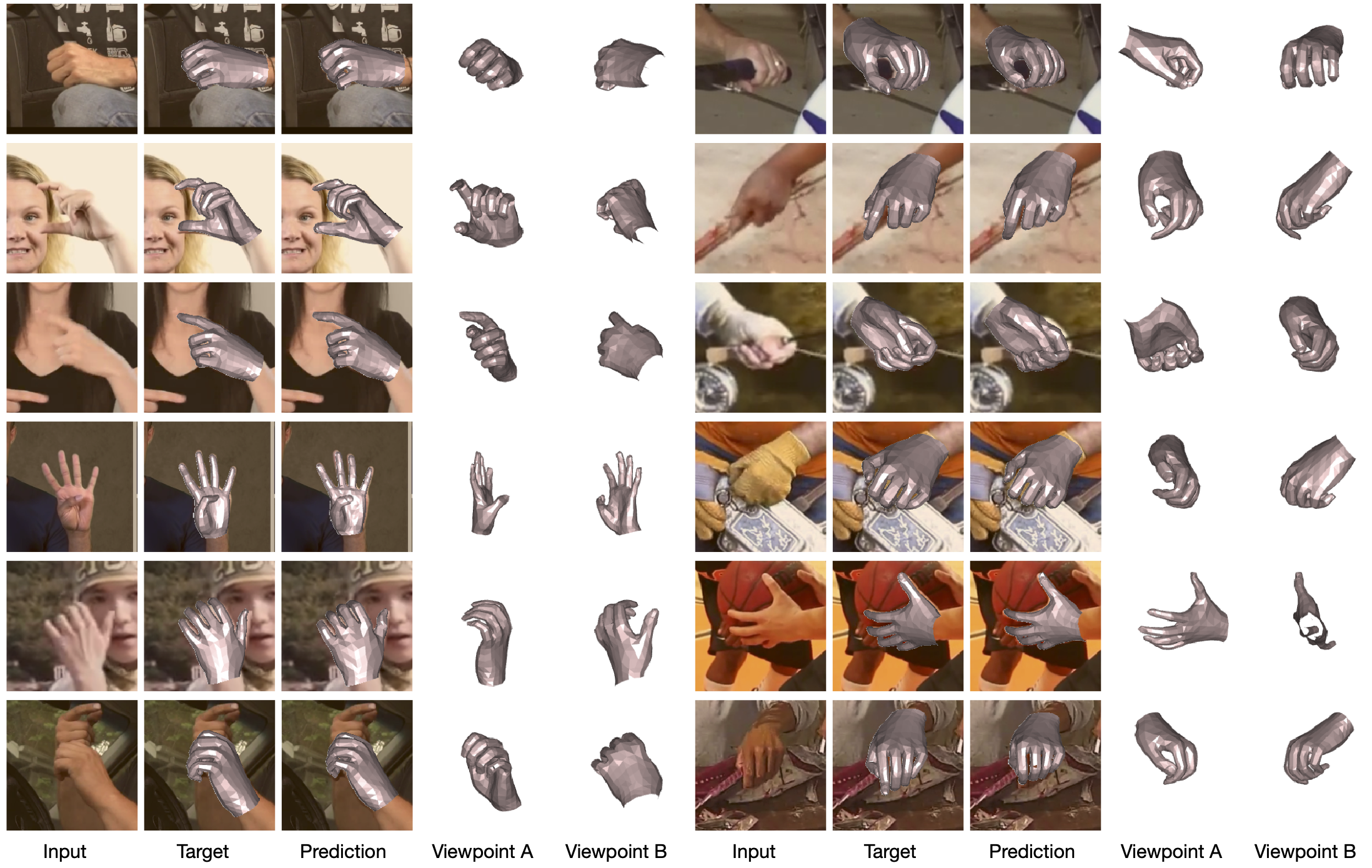}
    \caption{Qualitative mesh reconstruction results shown on \texttt{YouTube} (left half) and \texttt{MPII} (right half).}
    \label{fig:qualit}
\end{figure*}

Qualitative results on the in the wild datasets can be found in Figure~\ref{fig:qualit}.

\subsection{Self-Comparison}

Table~\ref{table:pose-self-eval} and Table~\ref{table:mesh-self-eval} show the pose estimation and mesh reconstruction error on different types of mesh decoders. To perform broad hyperparameter search that ensures fair comparison of the baseline methods, we use a reduced training dataset, corresponding to row 6 in  Table~\ref{table:dataset-eval}, to speed-up optimization.

\texttt{Spiral} refers to the proposed approach, \texttt{Spectral} follows the CoMA decoder~\cite{DBLP:journals/corr/abs-1807-10267} implemented as in Kulon et al.~\cite{dkulon2019rec} but with no camera estimation branch. \texttt{Spiral GMM} is the spiral implementation of \cite{dkulon2019rec} which incorporates decoder pretraining that is fixed afterwards and a camera estimation branch. \texttt{Spiral GMM, tuned} is a fine-tuned version obtained by resuming training of all parameters of the \texttt{Spiral GMM}. Finally, \texttt{MANO} is similar to Boukhayma et al.~\cite{2019arXiv190203451B} but with the loss function on mesh vertices instead of joints. We observe that our edge length preservation loss term stabilizes \texttt{MANO} shape parameters regression that in prior works was addressed by imposing a large L2 penalty~\cite{2019arXiv190203451B} to mitigate divergence.

We find that a spatial method outperforms a spectral approach. Moreover, end-to-end training with loss on meshes in the image coordinate system is better than pretraining in the canonical frame and estimating camera parameters.

\begin{table}[t]
  \centering
  \resizebox{\columnwidth}{!}{%
  \begin{tabular}{||c || c c c  || } 
 \hline
 Decoder Type & RHD [mm] & MPII [px] & YouTube [mm] \\ [0.5ex] 
 \hline\hline
 MANO & 12.086 & 11.448 & 14.958  \\ 
 \hline
 Spectral & 11.638 & 9.858 & 13.612  \\ 
 \hline
 Spiral GMM & 11.138 & 10.074 & 11.999  \\ 
 \hline
 Spiral GMM, tune & 11.121 & 10.117 & 11.799  \\ 
 \hline
 Spiral & \textbf{11.052} & \textbf{9.434} & \textbf{10.698}  \\ 
 \hline
\end{tabular}
}
  \caption{Pose Error. Performance with different decoders trained on a reduced dataset.}
\label{table:pose-self-eval}
\end{table}

\begin{table}[t]
  \centering
  \resizebox{0.8\columnwidth}{!}{%
  \begin{tabular}{||c || c c  || } 
 \hline
 Decoder Type & MAE [mm] & \begin{tabular}[c]{@{}c@{}} Mesh \\ Error {[}mm{]} \end{tabular} \\ [0.5ex] 
 \hline\hline
 MANO & 13.023 & 27.485  \\ 
 \hline
 Spectral & 9.058 & 19.317  \\ 
 \hline
 Spiral GMM & 8.462 & 18.206 \\ 
 \hline
 Spiral GMM, tune & 8.380 & 18.057  \\ 
 \hline
 Spiral & \textbf{7.827} & \textbf{17.096} \\ 
 \hline
\end{tabular}
}
  \caption{Mesh reconstruction with different decoders on the \texttt{YouTube} dataset trained on a reduced dataset.}
\label{table:mesh-self-eval}
\end{table}

\subsection{Comparison With Iterative Fitting}

\begin{table}[b]
  \centering
  \resizebox{0.8\columnwidth}{!}{
  \begin{tabular}{||c || c  c || } 
 \hline
 Model & MPII [px] & MPII Filtered [px]\\ [0.5ex] 
 \hline\hline
 Iterative Fitting & 10.295 & \textbf{5.111} \\ 
 \hline
 Network & \textbf{9.265} & 5.555 \\ 
 \hline
\end{tabular}
}
  \caption{Comparison of the network performance with iterative fitting on the \texttt{MPII} test set in terms of Pose Error.}
\label{table:net-vs-fitting}
\end{table}

Our system is supervised with meshes iteratively aligned with keypoint predictions. Naturally the question arises whether the network performance is bounded by the performance of iterative fitting. We compare both approaches on the only existing manually annotated in the wild dataset. Additionally, we select 390 out of 846 samples that were filtered according to sanity checks described in Section~\ref{section:automated}. 

Based on Table~\ref{table:net-vs-fitting} we observe that our system performs slightly worse to iterative fitting on a filtered dataset. This is expected as these are samples on which the keypoint estimator has very high confidence. Importantly, the network performs much better when the whole dataset is considered.

The network is also characterized by the strikingly faster inference time. Specifically, the reported timing (Section~\ref{section:fitting}) for iterative fitting of 4000 samples / 10 minutes is with a batch size of 4000. A single sample takes 110 seconds (GPU) or 70 seconds (CPU) for fitting to 2D annotations, exluding keypoint predictions. By contrast, our network inference time is 60 FPS (GeForce RTX 2080 Ti).

\begin{table}[b]
  \centering
  \resizebox{\columnwidth}{!}{
  \begin{tabular}{||c || c c c c c || c c c ||} 
 \hline
 ID & MPII & RHD & MTC & FreiHAND & YouTube & \begin{tabular}[c]{@{}c@{}} RHD \\ {[}mm{]} \end{tabular} & \begin{tabular}[c]{@{}c@{}} MPII \\ {[}px{]} \end{tabular} & \begin{tabular}[c]{@{}c@{}} YouTube \\ {[}mm{]} \end{tabular} \\ [0.5ex] 
 \hline\hline
 1 & X &  &  &   &  &  15.398 & 14.930 & 35.032 \\ 
 2 &  &  &   & X &  & 13.258  & 16.558 & 23.409 \\ 
 3 &  &  &   &  & X & 12.448 & 10.572 & 11.052 \\ 
 \hline \hline 
 4 & X & X &  &   &  &  11.205 & 11.547 & 18.539 \\
 \hline \hline
 5 & X & X & X &   &  & 11.202 & 10.965 & 16.214 \\ 
 6 & X & X & X &   & X & 11.052 & 9.434 & 10.698 \\ 
 7 & X & X & X &  X &  & 10.926 & 9.672 & 12.637 \\ 
 8 & X & X & X &  X & X & 10.922 & 9.292 & 10.536 \\
 \hline
\end{tabular}
}
\caption{Pose Error. Performance on different combinations of datasets.}
\label{table:dataset-eval}
\end{table}

\subsection{Dataset and Method Ablation}
In Table~\ref{table:dataset-eval}, we evaluate the system trained on different combinations of datasets. 
By comparing rows 1-3, we observe that training a mesh reconstruction system on our dataset leads to better results than training on other outdoor datasets. Rows 5-8, show that the best results are obtained when our dataset is included during training, even when adding it to the union of all other datasets. We also observe state-of-the-art performance on \texttt{RHD} and \texttt{MPII} without recent datasets such as \texttt{YouTube}, \texttt{FreiHAND}, \texttt{MTC} proving the efficiency of the proposed neural network (Row 4). Similarly,  Table~\ref{table:pose-self-eval} shows that the MANO-based method trained with the novel dataset and mesh loss obtains better performance than prior MANO-based implementations indicating the importance of the contributed data collection system.

\section{Conclusion}
We have shown that a simple encoder-decoder architecture can obtain superior performance on mesh reconstruction and pose estimation tasks given an appropriately defined objective function. Furthermore, we have proposed an approach for automated data collection that allows adapting the system to non-laboratory domains and additionally improves results on common benchmarks. 
We believe that these findings justify looking further into mesh generative models for human modeling, as well as other weakly- and self-supervised methods that alleviate the need of acquiring 3D ground truth for humans in unconstrained environments. 

\paragraph*{Acknowledgments}
M.B. was partially supported by the ERC Consolidator Grant No.~724228 (LEMAN). S.Z. was partially funded by the EPSRC Fellowship DEFORM (EP/S010203/1) and an Amazon AWS Machine Learning Research Award. D.K. was funded by a PhD scholarship.

\clearpage
%%%%%%% REFERENCES
{\small
\bibliographystyle{ieee_fullname}
\bibliography{egbib}

\begin{thebibliography}{10}\itemsep=-1pt

\bibitem{6909866}
M. Andriluka, L. Pishchulin, P. Gehler, and B. Schiele.
\newblock 2d human pose estimation: New benchmark and state of the art
  analysis.
\newblock In {\em 2014 IEEE Conference on Computer Vision and Pattern
  Recognition}, pages 3686--3693, June 2014.

\bibitem{2019arXiv190404196B}
Seungryul Baek, Kwang~In Kim, and Tae-Kyun Kim.
\newblock Pushing the envelope for rgb-based dense 3d hand pose estimation via
  neural rendering.
\newblock In {\em The IEEE Conference on Computer Vision and Pattern
  Recognition (CVPR)}, June 2019.

\bibitem{Bogo:ECCV:2016}
Federica Bogo, Angjoo Kanazawa, Christoph Lassner, Peter Gehler, Javier Romero,
  and Michael~J. Black.
\newblock Keep it {SMPL}: Automatic estimation of {3D} human pose and shape
  from a single image.
\newblock In {\em Computer Vision -- ECCV 2016}, Lecture Notes in Computer
  Science. Springer International Publishing, Oct. 2016.

\bibitem{DBLP:journals/corr/BoscainiMRB16}
Davide Boscaini, Jonathan Masci, Emanuele Rodol{\`a}, and Michael Bronstein.
\newblock Learning shape correspondence with anisotropic convolutional neural
  networks.
\newblock In {\em Advances in Neural Information Processing Systems}, pages
  3189--3197, 2016.

\bibitem{2019arXiv190203451B}
Adnane Boukhayma, Rodrigo~de Bem, and Philip~HS Torr.
\newblock 3d hand shape and pose from images in the wild.
\newblock In {\em Proceedings of the IEEE Conference on Computer Vision and
  Pattern Recognition}, pages 10843--10852, 2019.

\bibitem{2019arXiv190502876B}
Giorgos Bouritsas, Sergiy Bokhnyak, Stylianos Ploumpis, Michael Bronstein, and
  Stefanos Zafeiriou.
\newblock Neural 3d morphable models: Spiral convolutional networks for 3d
  shape representation learning and generation.
\newblock In {\em The IEEE International Conference on Computer Vision (ICCV)},
  2019.

\bibitem{bronstein2017geometric}
Michael~M Bronstein, Joan Bruna, Yann LeCun, Arthur Szlam, and Pierre
  Vandergheynst.
\newblock Geometric deep learning: going beyond euclidean data.
\newblock {\em IEEE Signal Processing Magazine}, 34(4):18--42, 2017.

\bibitem{DBLP:journals/corr/BrunaZSL13}
Joan Bruna, Wojciech Zaremba, Arthur Szlam, and Yann LeCun.
\newblock Spectral networks and locally connected networks on graphs.
\newblock In {\em 2nd International Conference on Learning Representations,
  {ICLR} 2014, Banff, AB, Canada, April 14-16, 2014, Conference Track
  Proceedings}, 2014.

\bibitem{cai2018_weakly}
Yujun Cai, Liuhao Ge, Jianfei Cai, and Junsong Yuan.
\newblock Weakly-supervised 3d hand pose estimation from monocular rgb images.
\newblock In {\em ECCV}, 2018.

\bibitem{cai2019exploiting}
Yujun Cai, Liuhao Ge, Jun Liu, Jianfei Cai, Tat-Jen Cham, Junsong Yuan, and
  Nadia~Magnenat Thalmann.
\newblock Exploiting spatial-temporal relationships for 3d pose estimation via
  graph convolutional networks.
\newblock In {\em The IEEE International Conference on Computer Vision (ICCV)},
  October 2019.

\bibitem{Freihand2019}
Jimei Yang Bryan Russell Max~Argus Christian~Zimmermann, Duygu~Ceylan and
  Thomas Brox.
\newblock Freihand: A dataset for markerless capture of hand pose and shape
  from single rgb images.
\newblock In {\em IEEE International Conference on Computer Vision (ICCV)},
  2019.

\bibitem{DBLP:journals/corr/DefferrardBV16}
Micha{\"e}l Defferrard, Xavier Bresson, and Pierre Vandergheynst.
\newblock Convolutional neural networks on graphs with fast localized spectral
  filtering.
\newblock In {\em Advances in neural information processing systems}, pages
  3844--3852, 2016.

\bibitem{Dibra_2018_CVPR_Workshops}
Endri Dibra, Silvan Melchior, Ali Balkis, Thomas Wolf, Cengiz Oztireli, and
  Markus Gross.
\newblock Monocular rgb hand pose inference from unsupervised refinable nets.
\newblock In {\em The IEEE Conference on Computer Vision and Pattern
  Recognition (CVPR) Workshops}, June 2018.

\bibitem{gao2019object}
Yafei Gao, Yida Wang, Pietro Falco, Nassir Navab, and Federico Tombari.
\newblock Variational object-aware 3d hand pose from a single rgb image.
\newblock {\em IEEE Robotics and Automation Letters}, PP:1--1, 07 2019.

\bibitem{garland1997surface}
Michael Garland and Paul~S Heckbert.
\newblock Surface simplification using quadric error metrics.
\newblock In {\em Proceedings of the 24th annual conference on Computer
  graphics and interactive techniques}, pages 209--216. ACM
  Press/Addison-Wesley Publishing Co., 1997.

\bibitem{ge2016robust}
Liuhao Ge, Hui Liang, Junsong Yuan, and Daniel Thalmann.
\newblock Robust 3d hand pose estimation in single depth images: from
  single-view cnn to multi-view cnns.
\newblock In {\em Proceedings of the IEEE conference on computer vision and
  pattern recognition}, pages 3593--3601, 2016.

\bibitem{2019arXiv190300812G}
Liuhao Ge, Zhou Ren, Yuncheng Li, Zehao Xue, Yingying Wang, Jianfei Cai, and
  Junsong Yuan.
\newblock 3d hand shape and pose estimation from a single rgb image.
\newblock In {\em Proceedings of the IEEE Conference on Computer Vision and
  Pattern Recognition}, pages 10833--10842, 2019.

\bibitem{gong2019spiralnet++}
Shunwang Gong, Lei Chen, Michael Bronstein, and Stefanos Zafeiriou.
\newblock Spiralnet++: A fast and highly efficient mesh convolution operator.
\newblock In {\em Proceedings of the IEEE International Conference on Computer
  Vision Workshops}, 2019.

\bibitem{Guler_2019_CVPR}
Riza~Alp Guler and Iasonas Kokkinos.
\newblock Holopose: Holistic 3d human reconstruction in-the-wild.
\newblock In {\em The IEEE Conference on Computer Vision and Pattern
  Recognition (CVPR)}, June 2019.

\bibitem{Hasson:CVPR:2019}
Yana Hasson, G{\"u}l Varol, Dimitrios Tzionas, Igor Kalevatykh, Michael~J.
  Black, Ivan Laptev, and Cordelia Schmid.
\newblock Learning joint reconstruction of hands and manipulated objects.
\newblock In {\em CVPR}, 2019.

\bibitem{DBLP:journals/corr/abs-1801-01615}
Hanbyul Joo, Tomas Simon, and Yaser Sheikh.
\newblock Total capture: A 3d deformation model for tracking faces, hands, and
  bodies.
\newblock In {\em Proceedings of the IEEE Conference on Computer Vision and
  Pattern Recognition}, pages 8320--8329, 2018.

\bibitem{hmr_kanazawa}
Angjoo Kanazawa, Michael~J Black, David~W Jacobs, and Jitendra Malik.
\newblock End-to-end recovery of human shape and pose.
\newblock In {\em Proceedings of the IEEE Conference on Computer Vision and
  Pattern Recognition}, pages 7122--7131, 2018.

\bibitem{kipf2017semi}
Thomas~N. Kipf and Max Welling.
\newblock Semi-supervised classification with graph convolutional networks.
\newblock In {\em International Conference on Learning Representations (ICLR)},
  2017.

\bibitem{Knapitsch:2017:TTB:3072959.3073599}
Arno Knapitsch, Jaesik Park, Qian-Yi Zhou, and Vladlen Koltun.
\newblock Tanks and temples: Benchmarking large-scale scene reconstruction.
\newblock {\em ACM Trans. Graph.}, 36(4):78:1--78:13, July 2017.

\bibitem{SPIN:ICCV:2019}
Nikos Kolotouros, Georgios Pavlakos, Michael~J. Black, and Kostas Daniilidis.
\newblock Learning to reconstruct {3D} human pose and shape via model-fitting
  in the loop.
\newblock In {\em International Conference on Computer Vision}, Oct. 2019.

\bibitem{Kolotouros_2019_CVPR}
Nikos Kolotouros, Georgios Pavlakos, and Kostas Daniilidis.
\newblock Convolutional mesh regression for single-image human shape
  reconstruction.
\newblock In {\em The IEEE Conference on Computer Vision and Pattern
  Recognition (CVPR)}, June 2019.

\bibitem{dkulon2019rec}
Dominik Kulon, Haoyang Wang, Riza~Alp G{\"{u}}ler, Michael~M. Bronstein, and
  Stefanos Zafeiriou.
\newblock Single image 3d hand reconstruction with mesh convolutions.
\newblock In {\em Proceedings of the British Machine Vision Conference
  ({BMVC})}, 2019.

\bibitem{DBLP:journals/corr/Lassner0KBBG17}
Christoph Lassner, Javier Romero, Martin Kiefel, Federica Bogo, Michael~J
  Black, and Peter~V Gehler.
\newblock Unite the people: Closing the loop between 3d and 2d human
  representations.
\newblock In {\em Proceedings of the IEEE Conference on Computer Vision and
  Pattern Recognition}, pages 6050--6059, 2017.

\bibitem{DBLP:journals/corr/LevieMBB17}
Ron Levie, Federico Monti, Xavier Bresson, and Michael~M Bronstein.
\newblock Cayleynets: Graph convolutional neural networks with complex rational
  spectral filters.
\newblock {\em IEEE Transactions on Signal Processing}, 67(1):97--109, 2018.

\bibitem{DBLP:journals/corr/abs-1809-06664}
Isaak Lim, Alexander Dielen, Marcel Campen, and Leif Kobbelt.
\newblock A simple approach to intrinsic correspondence learning on
  unstructured 3d meshes.
\newblock In {\em Proceedings of the European Conference on Computer Vision
  (ECCV)}, pages 0--0, 2018.

\bibitem{2014arXiv1405.0312L}
Tsung-Yi Lin, Michael Maire, Serge Belongie, James Hays, Pietro Perona, Deva
  Ramanan, Piotr Doll{\'a}r, and C~Lawrence Zitnick.
\newblock Microsoft coco: Common objects in context.
\newblock In {\em European conference on computer vision}, pages 740--755.
  Springer, 2014.

\bibitem{8621059}
J. {Liu}, H. {Ding}, A. {Shahroudy}, L. {Duan}, X. {Jiang}, G. {Wang}, and A.
  {Kot Chichung}.
\newblock Feature boosting network for 3d pose estimation.
\newblock {\em IEEE Transactions on Pattern Analysis and Machine Intelligence},
  pages 1--1, 2019.

\bibitem{SMPL:2015}
Matthew Loper, Naureen Mahmood, Javier Romero, Gerard Pons-Moll, and Michael~J.
  Black.
\newblock {SMPL}: A skinned multi-person linear model.
\newblock {\em ACM Trans. Graphics (Proc. SIGGRAPH Asia)}, 34(6):248:1--248:16,
  Oct. 2015.

\bibitem{DBLP:journals/corr/MasciBBV15}
Jonathan Masci, Davide Boscaini, Michael Bronstein, and Pierre Vandergheynst.
\newblock Geodesic convolutional neural networks on riemannian manifolds.
\newblock In {\em Proceedings of the IEEE international conference on computer
  vision workshops}, pages 37--45, 2015.

\bibitem{nzsl}
R. McKee, D. McKee, D. Alexander, and E. Paillat.
\newblock Nz sign language exercises.
\newblock Deaf Studies Department of Victoria University of Wellington.

\bibitem{DBLP:journals/corr/MontiBMRSB16}
Federico Monti, Davide Boscaini, Jonathan Masci, Emanuele Rodola, Jan Svoboda,
  and Michael~M Bronstein.
\newblock Geometric deep learning on graphs and manifolds using mixture model
  cnns.
\newblock In {\em Proceedings of the IEEE Conference on Computer Vision and
  Pattern Recognition}, pages 5115--5124, 2017.

\bibitem{2017arXiv171201057M}
Franziska Mueller, Florian Bernard, Oleksandr Sotnychenko, Dushyant Mehta,
  Srinath Sridhar, Dan Casas, and Christian Theobalt.
\newblock Ganerated hands for real-time 3d hand tracking from monocular rgb.
\newblock In {\em Proceedings of the IEEE Conference on Computer Vision and
  Pattern Recognition}, pages 49--59, 2018.

\bibitem{oberweger2017deepprior++}
Markus Oberweger and Vincent Lepetit.
\newblock Deepprior++: Improving fast and accurate 3d hand pose estimation.
\newblock In {\em Proceedings of the IEEE International Conference on Computer
  Vision}, pages 585--594, 2017.

\bibitem{DBLP:journals/corr/abs-1808-05942}
Mohamed Omran, Christoph Lassner, Gerard Pons-Moll, Peter Gehler, and Bernt
  Schiele.
\newblock Neural body fitting: Unifying deep learning and model based human
  pose and shape estimation.
\newblock In {\em 2018 International Conference on 3D Vision (3DV)}, pages
  484--494. IEEE, 2018.

\bibitem{2017arXiv171203866P}
Paschalis Panteleris, Iason Oikonomidis, and Antonis Argyros.
\newblock Using a single rgb frame for real time 3d hand pose estimation in the
  wild.
\newblock In {\em 2018 IEEE Winter Conference on Applications of Computer
  Vision (WACV)}, pages 436--445. IEEE, 2018.

\bibitem{SMPL-X:2019}
Georgios Pavlakos, Vasileios Choutas, Nima Ghorbani, Timo Bolkart, Ahmed A.~A.
  Osman, Dimitrios Tzionas, and Michael~J. Black.
\newblock Expressive body capture: 3d hands, face, and body from a single
  image.
\newblock In {\em Proceedings IEEE Conf. on Computer Vision and Pattern
  Recognition (CVPR)}, 2019.

\bibitem{2018arXiv180504092P}
Georgios Pavlakos, Luyang Zhu, Xiaowei Zhou, and Kostas Daniilidis.
\newblock Learning to estimate 3d human pose and shape from a single color
  image.
\newblock In {\em Proceedings of the IEEE Conference on Computer Vision and
  Pattern Recognition}, pages 459--468, 2018.

\bibitem{DBLP:journals/corr/abs-1807-10267}
Anurag Ranjan, Timo Bolkart, Soubhik Sanyal, and Michael~J Black.
\newblock Generating 3d faces using convolutional mesh autoencoders.
\newblock In {\em Proceedings of the European Conference on Computer Vision
  (ECCV)}, pages 704--720, 2018.

\bibitem{MANO:SIGGRAPHASIA:2017}
Javier Romero, Dimitrios Tzionas, and Michael~J. Black.
\newblock Embodied hands: Modeling and capturing hands and bodies together.
\newblock {\em ACM Transactions on Graphics, (Proc. SIGGRAPH Asia)},
  36(6):245:1--245:17, Nov. 2017.
\newblock (*) Two first authors contributed equally.

\bibitem{DBLP:journals/corr/SimonJMS17}
Tomas Simon, Hanbyul Joo, Iain Matthews, and Yaser Sheikh.
\newblock Hand keypoint detection in single images using multiview
  bootstrapping.
\newblock In {\em Proceedings of the IEEE conference on Computer Vision and
  Pattern Recognition}, pages 1145--1153, 2017.

\bibitem{spurr2018cvpr}
Adrian Spurr, Jie Song, Seonwook Park, and Otmar Hilliges.
\newblock Cross-modal deep variational hand pose estimation.
\newblock In {\em CVPR}, 2018.

\bibitem{DBLP:journals/corr/SupancicRYSR15}
James~S Supancic, Gr{\'e}gory Rogez, Yi Yang, Jamie Shotton, and Deva Ramanan.
\newblock Depth-based hand pose estimation: data, methods, and challenges.
\newblock In {\em Proceedings of the IEEE international conference on computer
  vision}, pages 1868--1876, 2015.

\bibitem{tan2017indirect}
Vince Tan, Ignas Budvytis, and Roberto Cipolla.
\newblock Indirect deep structured learning for 3d human body shape and pose
  prediction.
\newblock {\em BMVC}, 2017.

\bibitem{tekin2019h+}
Bugra Tekin, Federica Bogo, and Marc Pollefeys.
\newblock H+ o: Unified egocentric recognition of 3d hand-object poses and
  interactions.
\newblock In {\em Proceedings of the IEEE Conference on Computer Vision and
  Pattern Recognition}, pages 4511--4520, 2019.

\bibitem{2018arXiv180404875V}
Gul Varol, Duygu Ceylan, Bryan Russell, Jimei Yang, Ersin Yumer, Ivan Laptev,
  and Cordelia Schmid.
\newblock Bodynet: Volumetric inference of 3d human body shapes.
\newblock In {\em Proceedings of the European Conference on Computer Vision
  (ECCV)}, pages 20--36, 2018.

\bibitem{velikovi2017graph}
Petar Veli{\v{c}}kovi{\'{c}}, Guillem Cucurull, Arantxa Casanova, Adriana
  Romero, Pietro Li{\`{o}}, and Yoshua Bengio.
\newblock Graph attention networks.
\newblock {\em International Conference on Learning Representations}, 2018.

\bibitem{DBLP:journals/corr/VermaBV17}
Nitika Verma, Edmond Boyer, and Jakob Verbeek.
\newblock Feastnet: Feature-steered graph convolutions for 3d shape analysis.
\newblock In {\em Proceedings of the IEEE Conference on Computer Vision and
  Pattern Recognition}, pages 2598--2606, 2018.

\bibitem{DBLP:journals/corr/WeiRKS16}
Shih-En Wei, Varun Ramakrishna, Takeo Kanade, and Yaser Sheikh.
\newblock Convolutional pose machines.
\newblock In {\em Proceedings of the IEEE Conference on Computer Vision and
  Pattern Recognition}, pages 4724--4732, 2016.

\bibitem{WetzlerBMVC15}
Aaron Wetzler, Ron Slossberg, and Ron Kimmel.
\newblock Rule of thumb: Deep derotation for improved fingertip detection.
\newblock In {\em Proceedings of the British Machine Vision Conference
  ({BMVC})}, 2015.

\bibitem{DBLP:journals/corr/abs-1812-01598}
Donglai Xiang, Hanbyul Joo, and Yaser Sheikh.
\newblock Monocular total capture: Posing face, body, and hands in the wild.
\newblock In {\em Proceedings of the IEEE Conference on Computer Vision and
  Pattern Recognition}, pages 10965--10974, 2019.

\bibitem{yang2019iccv}
Linlin Yang, Shile Li, Dongheui Lee, and Angela Yao.
\newblock Aligning latent spaces for 3d hand pose estimation.
\newblock In {\em The IEEE International Conference on Computer Vision (ICCV)},
  October 2019.

\bibitem{yang2018disentangling}
Linlin Yang and Angela Yao.
\newblock Disentangling latent hands for image synthesis and pose estimation.
\newblock In {\em Proceedings of the IEEE Conference on Computer Vision and
  Pattern Recognition}, pages 9877--9886, 2019.

\bibitem{DBLP:journals/corr/abs-1712-03917}
Shanxin Yuan, Guillermo Garcia{-}Hernando, Bj{\"{o}}rn Stenger, Gyeongsik Moon,
  Ju~Yong Chang, Kyoung~Mu Lee, Pavlo Molchanov, Jan Kautz, Sina Honari, Liuhao
  Ge, Junsong Yuan, Xinghao Chen, Guijin Wang, Fan Yang, Kai Akiyama, Yang Wu,
  Qingfu Wan, Meysam Madadi, Sergio Escalera, Shile Li, Dongheui Lee, Iason
  Oikonomidis, Antonis~A. Argyros, and Tae{-}Kyun Kim.
\newblock Depth-based 3d hand pose estimation: From current achievements to
  future goals.
\newblock In {\em Proceedings of the IEEE Conference on Computer Vision and
  Pattern Recognition}, pages 2636--2645, 2018.

\bibitem{DBLP:journals/corr/ZhangJCQXY16}
J. {Zhang}, J. {Jiao}, M. {Chen}, L. {Qu}, X. {Xu}, and Q. {Yang}.
\newblock A hand pose tracking benchmark from stereo matching.
\newblock In {\em 2017 IEEE International Conference on Image Processing
  (ICIP)}, pages 982--986, Sep. 2017.

\bibitem{2019arXiv190209305Z}
Xiong Zhang, Qiang Li, Hong Mo, Wenbo Zhang, and Wen Zheng.
\newblock End-to-end hand mesh recovery from a monocular rgb image.
\newblock In {\em Proceedings of the IEEE International Conference on Computer
  Vision}, pages 2354--2364, 2019.

\bibitem{DBLP:journals/corr/ZhuPIE17}
Jun-Yan Zhu, Taesung Park, Phillip Isola, and Alexei~A Efros.
\newblock Unpaired image-to-image translation using cycle-consistent
  adversarial networks.
\newblock In {\em Proceedings of the IEEE international conference on computer
  vision}, pages 2223--2232, 2017.

\bibitem{DBLP:journals/corr/ZimmermannB17}
Christian Zimmermann and Thomas Brox.
\newblock Learning to estimate 3d hand pose from single rgb images.
\newblock In {\em IEEE International Conference on Computer Vision (ICCV)},
  2017.

\end{thebibliography}
}

\end{document}